\documentclass{article}
\usepackage{spconf,amsmath,graphicx}
\usepackage{comment}
\usepackage{color}
\specialcomment{notes}{\begingroup\sffamily\color{blue}}{\endgroup}
\usepackage{amsmath}
\usepackage{amssymb}

\usepackage{multirow}
\usepackage{subcaption}

\def\seqW{{\overrightarrow{w}}}
\def\arcW{{c}}

\usepackage{fancyhdr}
 
\fancypagestyle{specialfooter}{%
  \fancyhf{}
  \lhead{Copyright 2018 IEEE}
\rhead{Accepted at the IEEE 2018 Spoken Language Technology (SLT 2018)}

\lfoot{\small \copyright 2018 IEEE. Personal use of this material is permitted. Permission from IEEE must be obtained for all other uses, in any current or future media, including reprinting/republishing this material for advertising or promotional purposes, creating new collective works, for resale or redistribution to servers or lists, or reuse of any copyrighted component of this work in other works}
}

\title{Scalable language model adaptation for spoken dialogue systems}
\name{Ankur Gandhe, Ariya Rastrow, Bjorn Hoffmeister}
\address{\tt{\{aggandhe,arastrow,bjornh\}@amazon.com}}

\excludecomment{notes}

\begin{document}
\thispagestyle{specialfooter}

%
\maketitle
\begin{abstract}
Language models (LM) for interactive speech recognition systems are trained on large amounts of data and the model parameters are optimized on past user data. New application intents and interaction types are released for these systems over time, imposing challenges to adapt the LMs since the existing training data is no longer sufficient to model the future user interactions. It is unclear how to adapt LMs to new application intents without degrading the performance on existing applications. In this paper, we propose a solution to (a) estimate n-gram counts directly from the hand-written grammar for training LMs and (b) use constrained optimization to optimize the system parameters for future use cases, while not degrading the performance on past usage.  We evaluated our approach on new applications intents for a personal assistant system and find that the adaptation improves the word error rate by up to 15\% on new applications even when there is no adaptation data available for an application. 
\end{abstract}
%

\section{Introduction}
\label{sec:intro}
The quality of the data used for language modeling (LM) directly determines the accuracy of an automatic speech recognition (ASR) system. Most commercial ASR systems use n-gram based LMs trained on large amounts of text data. In spoken dialog systems like \textit{Amazon Alexa} and \textit{Google Home}, transcribed user utterances, collected from live traffic, form a large part of the training data. The transcribed data is also used for optimizing the LM parameters. Due to this, the data for training and tuning the LM parameters is always from the past user usage. When a new application intent is added to the system\footnote{An example of a new application intent is the ability to ask for cooking recipes}, the past user data is no longer sufficient to model the user interactions expected with the release of the application intent. This mismatch often leads to a reduced recognition quality on recently launched applications. 

In the absence of text data for training n-gram models, a common approach is to use a probabilistic context-free grammar (pCFG), similar to one described in~\cite{jurafsky1995using}. Figure~\ref{fig:grammar} shows an example grammar for a new application intent ``cooking recipes". The grammar is written to encompass the expected user interaction for the new application. The non terminals (\textit{DISH\_NAME} in the example) are expanded by specifying a catalog of entities for each non-terminal. Further, weights can be specified for the phrases and entities in the grammar to assign a probability to each sentence generated by the pCFG. In this sense, the pCFG can be regarded as a LM which is specific to a particular application intent. 

 \begin{figure}
\small
 \begin{verbatim}
i_want_to = (("[i]" ("[want]" | "[need"] | 
			            ("[would]" "[like"])) "[to"]));
action = ("[prepare]" | 
								         	"[cook]" | "[bake]" );
food_or_drink = (["food"] | DISH_NAME);
cook_dish = ( i_want_to action food_or_drink)
\end{verbatim}
\normalsize
\caption{Example grammar written for an application intent "cooking recipe'' in thrax~\cite{thrax} format}
\label{fig:grammar}
\end{figure}
The LM of an ASR system can be adapted to the new application by combining the pCFG with an existing model. A common approach is to sample data from the grammar~\cite{galescu1998rapid},~\cite{weilhammer2006bootstrapping}, or transform out-of-domain sentences into in-domain sentences using a simulator~\cite{chung2005automatic} and combine this data with pre-existing training data. However, certain grammars, e.g. those with large non-terminal, can encode exponentially many different paths, and hence, a large amount of data must be generated from these grammars to capture all possible variations. Another approach is to represent the final LM as a union of the grammar and n-gram models, where each of them is represented as an FST~\cite{rastrow2016speech},~\cite{mohri2002weighted}. However, rapidly accelerating spoken dialogue systems introduce ten's of new applications every month, and the final model can become quite large and lead to increased latency. In section~\ref{sec:sampling_from_grammar}, we introduce a novel method for extracting n-gram counts directly from the grammar, thereby eliminating the need for generating sampled data.  These counts can be then used to train a n-gram based model or as additional data for training neural network based models. 


Language model adaptation has also been studied in recent studies. ~\cite{park2010improved} adapted a recurrent neural network LM to a single domain while ~\cite{alumae2013multi}, ~\cite{tilk2014multi} built a multi-domain neural network model by using domain vectors to modify the model output. However, an important missing aspect of the aforementioned approaches is the ways of optimizing the LM training for multiple domains while maintaining the performance on past usage.  In section~\ref{sec:constrained_optimization}, we introduce a constrained optimization criteria that minimizes the loss on a new application intents while not degrading the performance on existing applications supported by the spoken dialogue system. Our work can be extended to adapting neural language models by applying the same constrained loss during network training. 

\begin{notes}
[Write about extending ASR systems with new application,  lack of data (add citation), add citation to grammar writing, catalog ingestion. Give example grammar and application that is added] 
\end{notes}
\begin{notes}
[related work]
~\cite{jurafsky1995using} describes how to use a grammar as an LM for ASR.
~\cite{galescu1998rapid} generate artificial corpora by generating sentences from a context-free grammar (CFG), also select  data from  OOD corpora. 
~\cite{wang2000unified} combine a CFG with an n-gram model, which is similar to class-based models. 
~\cite{chung2005automatic} use a simulator to generate a in-domain seed corpus and use it to transform out-of-domain sentences into in-domain sentences. Requires a good simulator of user utterances. 
~\cite{georges2013transducer} also works by sampling from grammars, generalization of a class-based LM.
~\cite{di2004bootstrapping} bootstrap from transcriptions from other domains.
~\cite{hakkani2006bootstrapping}   bootstrap from the world wide web.
~\cite{weilhammer2006bootstrapping} show that mixing data from grammar and general corpus can improve. Dont show how much to interpolate and optimize on multiple domains. 
~\cite{park2010improved} worked on adapting RNNLM to a single domain. Our work can be extended to RNNLM by applying the same constrained loss to during NN training. 
~\cite{alumae2013multi}, ~\cite{tilk2014multi} build a multi-domain RNNLM by using domain vectors to modify the model output while ~\cite{mikolov2012context} use a topic input to the RNNLM, but unlike our case, the topic of the input utterance is known at the time of decoding. 

\end{notes}

\section{Estimating n-gram models from pCFG}
\label{sec:sampling_from_grammar}
A language model estimates the probability of a given sequence of words $\seqW$. N-gram LMs make the assumption that the probability of word $w_i$ depends only on previous $n-1$ words, so that the probability can be written as:
\begin{equation}
\begin{aligned}
P(\seqW) = \prod_{|\seqW|} p(w_i| w_{i-1}, w_{i-2}, ... w_{i-n+1}) \\
\end{aligned}
\end{equation} 
For maximum likelihood estimation, it can be shown that $p(w_i|h) = \frac{C (w_i, h)}{\sum_j C(w_j, h)}$ where $h = w_{i-1}, w_{i-2}, ... w_{i-n+1}$ and $C(w_i, h)$ is count of the n-gram $w_i, h$ in the training data. Furthermore, for most commonly used smoothing techniques~\cite{chen1999empirical}, n-gram counts are sufficient to estimate the n-gram LM. In this paper, we extract fractional n-gram counts directly from a pCFG using the algorithm described in~\ref{sec:expected_counts}, and use smoothing techniques that can work with expected counts to estimate the n-gram probabilities.  
\subsection{Expected n-gram counts}
\label{sec:expected_counts}
A pCFG can be represented as a weighted finite state transducer (FST) where each arc represents either a word or a non-terminal, and the weights on the arcs are the probabilities associated with them; each non-terminal is also represented as an FST. When a sentence is sampled from the FST, the non-terminal arcs are replaced by the non-terminal FST to generate the final sentence.  \\
Each sequence $\seqW$ generated from the FST has an associated probability p($\seqW$). Then, the expected count of an n-gram $w_1, w_2, ... w_n$ within this sequence is same as  $p(\seqW)$. The expected count of an n-gram for the entire FST can be calculated by generating all possible sequences and summing their expected counts:  
\begin{equation}
\mathbb{E}(C(w_1, ... w_n))  = \sum_{\seqW_i \in FST} p(\seqW_i)(C_i(w_1, ... w_n) \in \seqW_i) 
\label{eq:expected-count}
\end{equation}
where $C_i(w_1,...,w_n)$ is the count of $w_1,...,w_n$ in $\seqW_i$. Computing this sum is non-trivial, as it requires enumerating all the sequences $\seqW_i$ in FST and calculating the normalized probability $p(\seqW_i)$. However, we can use dynamic programming to efficiently compute the n-gram counts as well as the normalized probability for each n-gram. \\

Each n-gram $w_1, ..., w_n$ is represented as a sequence of arcs $arc_1, .. arc_n$ in an FST and the expected count of the n-gram can be computed by the sum of the probability of all paths that include the corresponding sequence of arcs.  If $s_1$ is the initial state of $arc_1$, $s_n$ is the final state of $arc_n$, then the sum of probabilities for the sequence of arcs is: 
\begin{equation}
p(arc_{1 \rightarrow n}) = \alpha_{s_1} * \arcW_{arc_1} * \arcW_{arc_2} ... * \arcW_{arc_n} * \beta_{s_n} / Z
\label{eq:arc_sum}
\end{equation}
where $\alpha_{s_1}$ and $\beta_{s_n}$ is the unnormalized sum of the probabilities of the paths ending in state $s_1$ and starting in  $s_n$ respectively, $\arcW_{arc_i}$ is the unnormalized probability of arc $arc_i$, and Z is the normalization factor ( sum of all paths in the FST). The values of $\alpha_s$, $\beta_s$ for every state in the FST and the normalization factor $Z$ can be computed efficiently using the forward-backward algorithm~\cite{rabiner1989tutorial}.  Then, the expected count of the n-gram is same as $p(arc_{1 \rightarrow n})$.\\

To compute the counts of all n-grams in the FST, we traverse the FST in a forward topological order. For each state, we iterate over all the incoming n-grams $w_1, w_2... w_{n-1}$ with their corresponding accumulated probability $p(arc_{1 \rightarrow n-1})$. For each out-going arc at the state, with word $w_n$, we calculate the expected count $\mathbb{E}(w_1, w_2, ..., w_n)$ using the following iterative equation:
\begin{equation}
p(arc_{1 \rightarrow n}) = p(arc_{1 \rightarrow n-1}) * \arcW_{arc_n} * \beta_{s_n} / \beta_{s_{n-1}}
\label{eq:arc_sum}
\end{equation}
and propagate the n-gram $w_1, w_2, ... w_n$ to the next state of the arc\footnote{Iterating over all n-grams can be expensive for adjacent non-terminals or cyclic grammars; this can be mitigated by applying heuristics such as not propagating forward n-grams with counts lower than a given threshold} along with its accumulated probability $p(arc_{1 \rightarrow n})$.

\subsection{Estimating Language Models}
Given the fractional counts extracted from a pCFG, smoothing algorithms that use fractional n-gram counts~\cite{zhang2014kneser},~\cite{kuznetsov2016learning} to build the n-gram model can be used.  Using count-of-count methods like Katz smoothing~\cite{katz1987estimation} and mod-KN smoothing~\cite{chen1999empirical} requires extracting count-of-counts from the pCFG, which can be done in a way similar to extracting fractional n-gram counts. Similarly, the n-gram counts can be used to train a feed-forward neural network or interpolate with an existing neural network model~\cite{neubig2016generalizing}. 
However, for simplicity, in this paper, we use Katz smoothing on scaled fractional counts to build the final n-gram model.   
\section{Optimizing interpolation weights}
\label{sec:constrained_optimization}
A pre-existing LM can be adapted to the new application intent by linear interpolation with the LM from section~\ref{sec:sampling_from_grammar}, where the probability of a word sequence $\seqW$ is calculated by using different interpolation weights, $\lambda_i$'s, for each LM in the mixture: 
\begin{equation}
\label{eq:interpolation}
p_{mix}(w_i|h_i) = \sum_{k} \lambda_k  p_k(w_i|h_i) 
\end{equation}
such that $\sum_k \lambda_k = 1$. In this approach, the interpolation weights are estimated by minimizing a loss, e.g. perplexity on representative data from the new application intent. When bootstrapping an existing LM with a grammar, however, just optimizing on the application's data is not sufficient; we need to make sure that the final LM does not degrade on existing applications. Hence, we propose a constrained optimization problem: 
\begin{equation}
\begin{aligned}
& \underset{\overrightarrow{\lambda}}{\text{minimize}}
& & Loss(D_{app}|\overrightarrow{\lambda}) \\
& \text{subject to}
& & Loss(D_{\text{past}}) \leq C
\end{aligned}
\label{eq:constrain_equation}
\end{equation}
where $\overrightarrow{\lambda} = \lambda_{app}, 1 - \lambda_{app}$ are interpolation weights for pre-existing LM and application intent LM respectively,  $Loss(D|\overrightarrow{\lambda})$ is the loss function being minimized, $D_{app}$ is representative data for the new application, and $D_{past}$ is development set based on past usage. 
\subsection{Loss functions}
\label{sec:loss_functions}
The actual loss function to be minimized (or constrained) depends on the adaptation data available for a new application intent. In this paper, we propose to use three loss functions:
\subsubsection{Negative squared minimization}
When no data is available for a application intent, the only way to optimize the LM for the new intent is to assign maximum possible interpolation weight to the feature LM. Hence, the loss can be expressed as:
\begin{equation}
Loss = - \lambda_{app}^2
\end{equation}
This will maximize the application intent's interpolation weight until we violate the constraint on past data.

\subsubsection{Perplexity minimization}
When sample text data, from the new application intent, is available for LM adaptation, one can minimize the perplexity of the interpolated model on the data
\begin{equation}
Loss = \exp ( - \frac{\sum_{i=1}^m log_e \, p_{mix}(w_i)}{m})
\end{equation}
where $p_{mix}(w_i)$ is the probability assigned by the interpolated model to the word $w_i$ and $m$ is the total number of words in the data.

\subsubsection{Expected WER minimization}
When there is transcribed audio available for an application intent, we can directly minimize the expected recognition error rate of the ASR system as explained in~\cite{liu2009use}. ~\cite{iyer1997analyzing} showed that perplexity is often not correlated with WER, and hence it might be better to directly minimize the WER when possible. Expected WER loss is computed by summing over the recognition error $\mathcal{E}
(\seqW|\seqW_{ref})$ for all hypotheses, $\seqW$s, weighted by their posterior probability $p(\seqW|O)$:
\begin{equation}
Loss = \sum_\seqW p(\seqW|O)\mathcal{E}(\seqW|\seqW_{ref}) 
\end{equation}
In order to make the sum over all possible hypothesis tractable, we approximate the sum by restricting $\seqW$s to an n-best list of ASR hypothesis.

\subsection{Constraint optimization}
The constraint on past usage is implemented as a penalty term~\cite{smith1997penalty} in the final optimization function: 
\begin{equation}
Loss_{constraint} = \sigma * max(0, Loss(D_{past} | \overrightarrow{\lambda}) - C))^2 
\end{equation}
where $C = Loss(D_{past} | Baseline)$, is the loss of the pre-existing LM, and $\sigma$ is the penalty coefficient to control the tolerance of the constraint. $\sigma$ can be set either statically to a high value or changed dynamically with every iteration of the optimization as suggested in ~\cite{yeniay2005penalty}. In this paper, we use a static penalty value of 1000. The loss function for the constraint can be either perplexity or expected WER. However, we expect that adding the constraint on WER will lead to better interpolation weights for the new application intent.
\subsection{Scaling to multiple applications} 
The proposed method can easily scale to adding multiple application intent grammars at the same time to an existing n-gram LM. We can extend equation~\ref{eq:constrain_equation} with one loss per application:
\[
Loss = \sum_{i=1}^N Loss_i(D_{app_i}|\overrightarrow{\lambda}) \\
\]
where $Loss_i$ is the loss function chosen for $app_i$ and $N$ is the total number of applications. The choice of loss function and the constraint used is independent of each application intent which enables us to choose the loss function for each intent depending on the availability of data for that application.
\begin{notes}
[Talk about L2 maximization, PPL minimization, expected nbestWER minimization, latticeWER minimization]
\end{notes}

\begin{figure*}[t]
\centering
\begin{subfigure}{.5\textwidth}
  \centering
  \includegraphics[width=.8\linewidth]{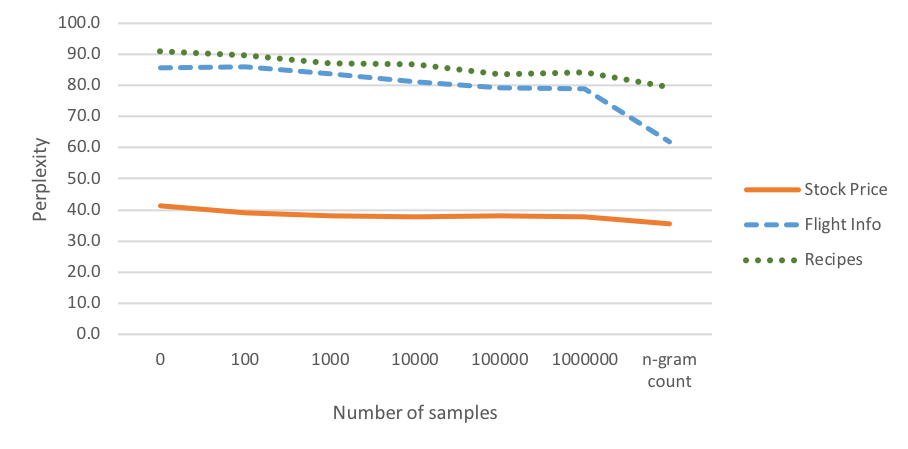}
  \caption{Perplexity}
  \label{fig:sub1}
\end{subfigure}%
\begin{subfigure}{.5\textwidth}
  \centering
  \includegraphics[width=0.8\linewidth]{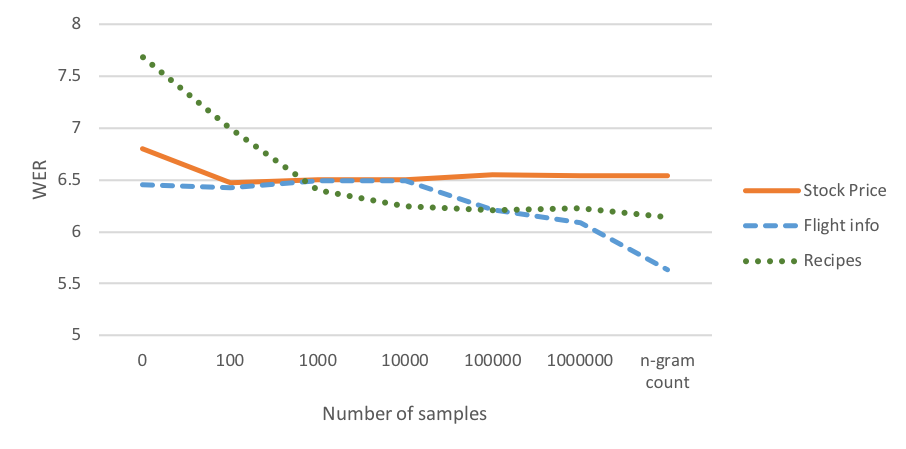}
    \caption{Word error rate}
  \label{fig:sub2}
\end{subfigure}
\caption{Figure above shows the perplexity and word error rate results when different amount of data is sampled from the grammars for generating n-gram models. Sampling size=0 shows the baseline perplexity and n-gram count is the method proposed in the paper}
\label{fig:sampling_vs_wer_ppl}
\end{figure*}


\begin{notes}
\begin{table*}[t]
\centering
\caption{Table below shows the perplexity results when we use different methods for generating n-gram models from grammar}
\label{tab:ngramcount}
\begin{tabular}{lllllllll}
            & \multicolumn{2}{c}{Past Data}     & \multicolumn{2}{c}{Stock Price}  & \multicolumn{2}{c}{Flights Info}    & \multicolumn{2}{c}{Recipes}     \\ \hline 
            & PPL                                                 & rel. WER & PPL                                                   & WER & PPL                                                  & WER & PPL     & WER \\ \hline
Baseline    & 32.4                                                &  -   & 41.6                                                  &   6.80  & 85.7                                                 &  7.42   & 90.9    &   7.68  \\
sample-sent & 34.2                                                &  0.5\%   & 37.5                                                  &   6.64  & 70.5                                                 &  7.42   & 83.5    &  6.83   \\
ngram-count & 34.5                                                &  1\%   & 33.8                                                  & 6.54    & 57.8                                                 &   7.26  & 79.2    &    6.14
\end{tabular}
\end{table*}
\end{notes}

\begin{table*}[]
\centering
\caption{Table showing improvement in perplexity and word error rate when minimizing different loss functions and different constraints}
\label{tab:constraint}
\begin{tabular}{llllllllll}
Constraint                                       & Loss function            & \multicolumn{2}{c}{Past Data}     & \multicolumn{2}{c}{Stock Price}  & \multicolumn{2}{c}{Flights Info}    & \multicolumn{2}{c}{Recipes}     \\ \hline 
                                                 &               & PPL           & rel. WER & PPL        & WER & PPL            & WER & PPL     & WER \\ \hline 
                                                  Baseline  &    &     32.4          &  -   & 41.6       &  6.80   & 85.6           &   6.45  & 90.9    &   7.68  \\ \hline \\ 
\multirow{3}{*}{PPL}                             & min L2        &      32.6        &  0\%   & 41.6       &  6.82   & 80.3          &  6.29  & 90.9    &  7.65   \\
                                                 & PPL           &        32.4       &   0\%  & 37.6       &   6.67  & 65.7           &   5.82  & 76.3    &    5.96 \\
                                                 & exp-WER   &        32.7       &  0\%   & 37.3       &  6.64   & 62.4           &  5.75   & 73.8    &   5.78  \\  \\ \hline \\ 
\multicolumn{1}{c}{\multirow{3}{*}{expWER}} & min L2        &     34.4          & 1\%   & 35.2       &  6.54   & 65.3           &  5.74   & 83.2    &   6.49  \\
\multicolumn{1}{c}{}                             & PPL           &      34.5         &   1\%  & 35.1       &  6.52   & 64.3           &   5.7  & 81      &  6.26   \\
\multicolumn{1}{c}{}                             & expWER   &       34.5        &    1\% & 35.1       &  6.54   & 61.9           &    5.63 & 79.2    &   6.14 
\end{tabular}
\end{table*}

\section{Experiments}
\label{sec:experiments}
The training data for the baseline language model is a combination of transcribed data from users of a personal assistant system as well as other in-house data sources\footnote{Data crawled from reddit.com, amazon forums, and amazon customer reviews}. We build one LM for each data source and interpolate them into a single LM. The interpolation weights are optimized on a held-out subset of transcribed data from users.

We evaluate our approach on grammars written for new application intents  which have little or no coverage in the LM training data. We tested with three application intents - \textit{Getting stock prices}, \textit{Getting Flight Information}, and \textit{Asking for recipes}. 

For running ASR experiments, we use an in-house ASR system\footnote{Trained on only a subset of data used for our production system} based on \cite{parthasarathi2015fmllr,garimella2015robust}. The acoustic model is a DNN trained on concatenated LFBE and i-vectors, and  the baseline LM is a 4-gram models with a vocabulary of 200K trained with modified kneser-ney smoothing~\cite{chen1999empirical}. The LMs for the data extracted from application grammars were trained using Katz smoothing, and we limited the number of new vocabulary words (new relative to baseline) to 10K. Each intent was tested for perplexity (PPL) and word error rate (WER) on a held out test set targeted towards the intent containing about 500 to 700 utterances\footnote{The test set was collected in-house by language experts and is not reflective of actual live usage of the application}. We also report the PPL and the relative WER degradation on past usage test set, dubbed as \textit{Past Data}.

\subsection{n-gram count extraction}
In section~\ref{sec:expected_counts}, we proposed extracting n-gram counts directly from the pCFG written for a new application intent. In figure~\ref{fig:sampling_vs_wer_ppl}, we compare the proposed method with a model trained on data sampled from the pCFGs on three application intents. The application LM is interpolated with the baseline model based on weights estimated using constrained optimization with expected word error as the loss and the constraint. We find that the difference in performance of the two techniques varies for different applications  - the perplexity difference is small for \textit{Recipes} and \textit{Stock Price} and significantly better for \textit{Flights info}. Similar results are observed on word error rate, although \textit{Stock Price} word error rate degrades with larger number of samples.  \\
We looked at  the total number of tri-grams (without replacing non-terminals) in the grammars for the applications intents as well as the total number of non-terminals in the three grammars, as shown in table~\ref{tab:fst_size}. \textit{Flight Info} and \textit{Stock Price} application grammars have much larger number of tri-grams in the grammar. However,  \textit{Flight Info} has a large number of non-terminals pairs (non-terminals appearing next to each other),  which leads to an exponentially large number of n-grams that can be sampled, and hence we observe a large drop in perplexity and word error rate when using the proposed method.  

\begin{table}
\begin{tabular}{llll}
            & Stock Price & Recipes & Flight Info \\ \hline
\# \small{of tri-grams} & 152k          & 49k      & 306k          \\ \hline
\# \small{of non-terminals} & 3          & 12      & 23        \\ \hline
\# \small{of non-terminal pair} & 3          & 23      & 203        \\ \hline

\end{tabular}
\caption{Number of paths and non-terminals in grammar of different intents}
\label{tab:fst_size}
\end{table}

\subsection{Comparing Loss function and constraint}
Section~\ref{sec:constrained_optimization} describes different methods for estimating interpolation weights, both for the loss to be minimized as well the constraint. Table~\ref{tab:constraint} shows the comparison of different loss functions and constraining on perplexity of baseline LM on the past data. It also compares the same loss functions but with expected WER as the constraint in the optimization. While both constraints ensure that the WER on past data degrades only marginally, constraint of expected WER leads to larger improvements in WER for new features. The improvements vary across the different application intent, and is more than 10\% relative for \textit{Flights Info} and \textit{Recipes} applications, which can be attributed to the initial model having a high perplexity on these applications. \\
We also find that L2 minimization works quite with expected WER constraint and the improvement in WER is comparable to other loss functions. This is quite useful as it shows that we can adapt to new application intents even when there is \textit{no adaptation data} available. The expected WER constraint uses the acoustic scores of each word when calculating the posterior probability, thereby reducing the dependency on the LM scores. This can explain why expected WER is less-stricter constraint than perplexity and assigns larger interpolation weights to new application intent LMs.

\subsection{Scalability}
\begin{figure}
\includegraphics[width=\linewidth]{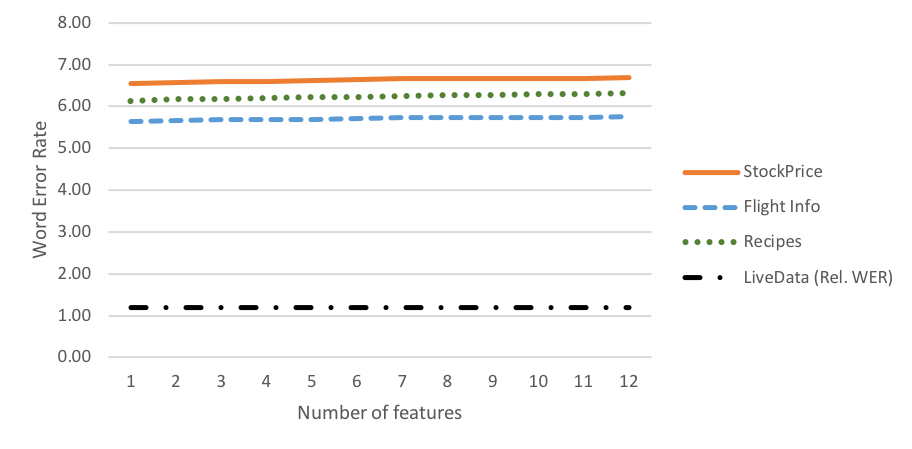}
\caption{Word error rate values of three different application intents and past data as more applications are added into the system} \label{fig:scalability}
\end{figure}

In the previous experiments, each application intent was optimized separately on its own loss function. However, our proposed method allows us to add as many new application intents into the optimization as we want. In figure~\ref{fig:scalability}, we show the word error rate change for the three  application intents as more and more application intents are added into the optimization\footnote{The new applications were chosen randomly from a set of held-out application intents. Some examples are \textit{Shopping}, \textit{Calendar events}, \textit{Donate to charity}.}. We find even though the word error rate increases with very new application added, the increase is not significant even with 12 new applications. 

\section{Conclusion}
\label{sec:conclusion}
We propose a new method for adapting the language model of a spoken dialogue system, specifically designed to scale to support multiple application intents. The proposed constrained optimization function ensures that the accuracy of the system does not degrade as new applications are added to the system. We also show that using word error rate as a constraint leads to better adaptation for the new applications and removes the need for any adaptation data. In the future, we want to extend the same framework for adaptation of neural network based LMs.

\bibliographystyle{IEEEbib}
\bibliography{mybib}

\end{document}